\documentclass[runningheads]{llncs}

\pdfoutput=1 
\usepackage[T1]{fontenc}
\usepackage{graphicx}

\usepackage[utf8]{inputenc}
\usepackage{textcmds}
\usepackage{lipsum}

\usepackage{graphicx}
\usepackage{wrapfig}

\usepackage{blindtext}
\usepackage{enumitem}
\usepackage{amsfonts}
\usepackage{amsmath}

\usepackage{tasks}
\usepackage{fancyref}

\usepackage{hyperref}

\usepackage{xcolor}
\definecolor{LightGray}{gray}{0.9}

\definecolor{codegreen}{rgb}{0,0.6,0}
\definecolor{codegray}{rgb}{0.5,0.5,0.5}
\definecolor{codepurple}{rgb}{0.58,0,0.82}
\definecolor{backcolour}{rgb}{0.95,0.95,0.92}

\begin{document}

\title{Parallel Neural Networks in Golang}

\author{Daniela Kalwarowskyj \and Erich Schikuta}

\institute{University of Vienna\\
Faculty of Computer Science, RG WST\\
A-1090 Vienna, Währingerstr. 29, Austria\\
\email{dkalwarowskyj@yahoo.com}\\
\email{erich.schikuta@univie.ac.at}}

\maketitle

\begin{abstract}
This paper describes the design and implementation of
parallel neural networks (PNNs) with the novel programming language Golang.
We follow in our approach the classical Single-Program Multiple-Data (SPMD) model where a PNN is composed of several sequential neural networks, which are
trained with a proportional share of the training dataset.
We used for this purpose the  MNIST dataset, which contains binary images of handwritten digits. Our analysis focusses on different activation functions
and optimizations in the form of stochastic gradients and
initialization of weights and biases.
We conduct a thorough performance analysis, where network configurations
and different performance factors are analyzed and interpreted.
Golang and its inherent parallelization support proved very well for parallel neural network simulation by considerable decreased processing times compared to sequential variants.

\keywords{Backpropagation Neuronal Network Simulation \and Parallel and Sequential Implementation \and MNIST \and Golang Programming Language}
\end{abstract}

\section{Introduction}
  When reading a letter our trained brain rarely has a problem to understand its meaning. Inspired by the way
  our nervous system perceives visual input, the idea emerged to write a mechanism that could \qq{learn} and
  furthermore use this \qq{knowledge} on unknown data. Learning is accomplished by repeating exercises and comparing
   results with given solutions.
  The neural network studied in this paper uses the MNIST dataset to train and test its capabilities.
  The actual learning is achieved by using backpropagation.
  In the course of our research, we concentrate on a single sequential feed
  forward neural network (SNN) and upgrade it into building multiple, parallel learning SNNs.
  Those parallel networks are then fused to one parallel neural network (PNN).
  These two types of networks are compared on their accuracy, confidence, computational performance and learning speed,
  which it takes those networks to learn the given task.

The specific contribution of the paper is twofold: on the one hand, a thorough analysis of sequential and parallel implementations of feed forward neural network respective time, accuracy and confidence, and on the other hand, a feasibility study of Golang~\cite{golang} and its tools for parallel simulation.

The structure of the paper is as follows:
In the next section, we give a short overview of related work. 
The parallelization approach is laid out in section~\ref{section:PNN} followed by the description of the Golang implementation. A comprehensive analysis of the sequential and parallel neural networks respective accuracy, confidence, computational performance and learning speed is presented in section~\ref{analysis}. Finally, the paper closes with a summary of the findings.

\section{Related Work and Baseline Research}
Artificial neural networks and their parallel simulation gained high attention in the scientific community.
Parallelization is a classic approach for speeding up execution times and exploiting the full potential of modern processors. Still, not every algorithm can profit from parallelization, as the concurrent execution might add a non-negligible overhead. This can also be the case for data parallel neural networks, where accuracy problems usually occur, as the results have to be merged.

In the literature a huge number of papers on parallelizing neural networks can be found. An excellent source of references is the survey by Tal Ben-Nun and Torsten Hoefler~\cite{ben2019demystifying}.
However, only few research was done on using Golang in this endeavour.

In the following only specific references are listed, which influenced the presented approach directly.
The authors of \cite{liu2016accuracy} presented a parallel backpropagation algorithm dealing with the accuracy problem only by using a MapReduce and Cascading model. 
In the course of our work on parallel and distributed systems~\cite{10.1007/BFb0057953,10.1007/3-540-62095-8_10,1380156} we developed several approaches for the parallelization of neural networks.
In \cite{schiki}, two novel parallel training approaches were presented for face recognizing backpropagation neural networks. The authors use the OpenMP environment for classic CPU multithreading and CUDA for parallelization on GPU architectures. Aside from that, they differentiated between topological data parallelism and structural data parallelism~\cite{schikuta1997structural}, where the latter is focus of the presented approach here. \cite{datavsmodel} gave a comparison of different parallelization approaches on a cluster computer. The results differed depending on the network size, data set sizes and number of processors. Besides parallelizing the backpropagation algorithm for training speed-up, alternative training algorithms like the Resilient Backpropagation described in \cite{rprop} might lead to faster convergence. One major difference to standard backpropagation is that every weight and bias has a different and variable learning rate. A detailed comparison of both network training algorithms was given in \cite{rpropcompare} in the case of spam classification.

%
%

\section{Fundamentals}\label{ann}
In the following we present the mathematical fundamentals of neural networks to allow for easier understanding and better applicability of our implementation approach described afterwards.


\subsubsection{Forwardpropagation}
To calculate an output in the last layer, the input values need to get propagated through each layer. This process is called forward propagation and is done by applying an activation function on each neuron's corresponding input sum.
The input sum $z$ for a neuron $k$ in the layer $l$ is the sum of each neuron's activation $a$ from the last layer multiplied with the weight $w$:
\begin{equation}
    z_{k}^{l}=\sum_{j}(w_{kj}^{l}a_{j}^{l-1}+b_{k}^{l})
\end{equation}
The additional term $+b$ stands for the bias value, which allows the activation function to be shifted to the left or to the right. For better readability, the input sums for a whole layer can be stored in a vector $z$ and defined by:
\begin{equation}
    z^{l}=W^{l}x^{l-1}+b^{l}
\end{equation}
Here, $W^l$ is a weight matrix storing all weights to layer $x^l$. To obtain the output of a layer, or, in case of the last layer $x^L$, the output of a neural network, an activation function $\varphi$ needs to be applied:
\begin{equation}
    x^{l}=\varphi(z^{l})=\varphi(W^{l}x^{l-1}+b^{l})
\end{equation}
Activation functions do not have to be unique in a network and can be combined. The implementation presented in this paper uses the rectifier activation function
\begin{equation}
    \varphi_{rectifier}(z)=
    \begin{cases}
        0       & \text{if } z < 0 \\
        z       & \text{if } z \geq 0
    \end{cases}
\end{equation}
for hidden neurons and the softmax activation function
\begin{equation}
   \varphi_{softmax}(z_i)=\frac{e^{z_i}}{\sum_{j}e^{z_j}}
\end{equation}
for output neurons. For classification, each class is represented by one neuron in the last layer. Due to the softmax function, the output values of those neurons sum up to 1 and can therefore be seen as the probabilities of being that class. 

\subsubsection{Backpropagation}
For proper classification the network has to be trained beforehand. In order to do that, a cost function tells us how well the network performs, like the cross entropy error with expected outputs $e$ and actual outputs $x$,
\begin{equation}
    C=-\sum_{i}e_{i}log(x_{i})
\end{equation}
The aim is to minimize the cost function by finding the optimal weights and biases with the gradient descent optimization algorithm. Therefore, a training instance gets forward propagated through the network to get an output. Subsequently, it is necessary to compute the partial derivatives of the cost function with respect to each weight and bias in the network:
\begin{equation}
    \frac{\partial{C}}{\partial{w_{kj}}}=\frac{\partial{C}}{\partial{z_{k}}}\frac{\partial{z_{k}}}{\partial{w_{kj}}}
\end{equation}
\begin{equation}
    \frac{\partial{C}}{\partial{b_{k}}}=\frac{\partial{C}}{\partial{z_{k}}}\frac{\partial{z_{k}}}{\partial{b_{kj}}}
\end{equation}
As a first step, $\frac{\partial{C}}{\partial{z_{k}}}$ needs to be calculated for every neuron $k$ in the last layer $L$:
\begin{equation}
    \delta_{k}^{L}=\frac{\partial{C}}{\partial{z_{k}^L}}=\frac{\partial{C}}{\partial{x_{k}^L}}\varphi'(z_{k}^L)
\end{equation}
In case of the cross entropy error function, the error signal vector $\delta$ of the softmax output layer is simply the actual output vector minus the expected output vector:
\begin{equation}
    \delta^{L}=\frac{\partial{C}}{\partial{z^L}}=x^{L}-e^{L}
\end{equation}
To obtain the errors for the remaining layers of the network, the output layer's error signal vector $\delta^L$ has to be propagated back through the network, hence the name of the algorithm:
\begin{equation}
   \delta^{l}=(W^{l+1})^{T}\delta^{l+1}\odot{\varphi'(z^l)}
\end{equation}
$(W^{l+1})^{T}$ is the transposed weight matrix, $\odot$ denotes the Hadamard product or entry-wise product and $\varphi'$ is the first derivative of the activation function.

\subsubsection{Gradient Descent}
Knowing the error of each neuron, the changes to the weights and biases can be determined by
\begin{equation}
    \Delta{w}_{kj}^{l}=-\eta\frac{\partial{C}}{\partial{w_{kj}^{l}}}=-\eta\delta_{k}^{l}x_{j}^{l-1}
\end{equation}
\begin{equation}
    \Delta{b}_{k}^{l}=-\eta\frac{\partial{C}}{\partial{b_{k}}}=-\eta\delta_{k}^{l}
\end{equation}
The constant $\eta$ is used to regulate the strength of the changes applied to the weights and biases and is also referred to as the learning rate, $x_j^{l-1}$ stands for the output of the $j^{th}$ neuron from layer $l-1$. The changes are applied by adding them to the old weights and biases.
Depending on the update frequency, a distinction is made between stochastic gradient descent, batch gradient descent and mini-batch gradient descent. In the case of the first-mentioned, the weights and biases are updated after every training instance (by repeating all of the aforementioned steps instance-wise). In contrast, batch gradient descent stands for updating only once after accumulating the gradients of all training samples. Mini-batch gradient descent is a combination of both. The weights and biases are updated after a specified amount, the \textit{mini-batch size}, of training instances. As with batch gradient descent, the gradients of all instances are averaged before the updates.

\section{Parallel Neuronal Networks}
\label{section:PNN}

This section describes the technology stack, the parallelization model and implementation details of the provided PNN.

\subsection{Technology Stack}

Go, often referred to as Golang, is a compiled, statically typed, open source programming language developed by a team at Google and released in November 2009. It is distributed under a BSD-style license, meaning that copying, modifying and redistributing is allowed under a few conditions.

As Andrew Gerrand, who works on the project, states in \cite{golang}, Go grew from a dissatisfaction with the development environments and languages that they were using at Google. It is designed to be expressive, concise, clean and efficient. Hence, Go compiles quickly and is as easy to read as it is to write. This is partly because of gofmt, the go source code formatter, that gives Go programmes a single style and relieves the programmers from discussions like where to set the braces. As uniform presentation makes code easier to read and therefore to work on, gofmt also saves time and affects the scalability of programming teams \cite{gokeynote}. The integrated garbage collector offers another great convenience and takes away the time consuming efforts on memory allocation and freeing known from C/C++. Despite the known overhead and criticism about Java's garbage collector, the author of \cite{gokeynote} claims that Go is different, more efficient and that it is almost essential for a concurrent language like Go because of the trickiness that can result from managing ownership of a piece of memory as it is passed around among concurrent executions. That being said, built-in support for concurrency is one of the most interesting aspects of Go, offering a great advantage over older languages like C++ or Java. One major component of Go's concurrency model are goroutines, which can be thought of as lightweight threads with a negligible overhead, as the cost of managing them is cheap compared to threads. If a goroutine blocks, the runtime automatically moves any blocking code away from being executed and executes some code that can run, leading to high-performance concurrency \cite{golang}. Communication between goroutines takes place over channels, which are derived from "Communicating Sequential Processes" found in \cite{csp}. A Channel can be used to send and receive messages from the type associated with it. Since receiving can only be done when something is being sent, channels can be used for synchronization, preventing race conditions by design.

Another difference to common object oriented programming languages can be found in Go's object oriented design. Its approach misses classes and type-based inheritance like subclassing, meaning that there is no type hierarchy. Instead, Go features polymorphism with interfaces and struct embedding and therefore encourages the composition over inheritance principle. An Interface is a set of methods, which is implemented implicitly by all data types that satisfy the interface \cite{gokeynote}.

For the rest, files are organized in packages, with every source file starting with a package statement. Packages can be used by importing them via their unique path. If a package path in the form of an URL refers to a remote repository, the remote package can be fetched with the \textit{go get} command and subsequently imported like a local package. Additionally, Go will not compile, if unused packages are being imported.

\subsection{Parallelization Model}

For the parallelization of neural network operations we apply the classical Single-Program Multiple-Data (SPMD) approach well known from high-performance computing~\cite{darema2001spmd}. It is a programming technique, where several tasks execute the same program but with different input data and the calculated output data is merged to a common result. Thus, based on the fundamentals of single feed forward neural network we generate multiple of these networks and set them up to work together in parallel manner.

\begin{figure}[ht]
  \begin{center}
\centering\includegraphics[width=0.6\linewidth]{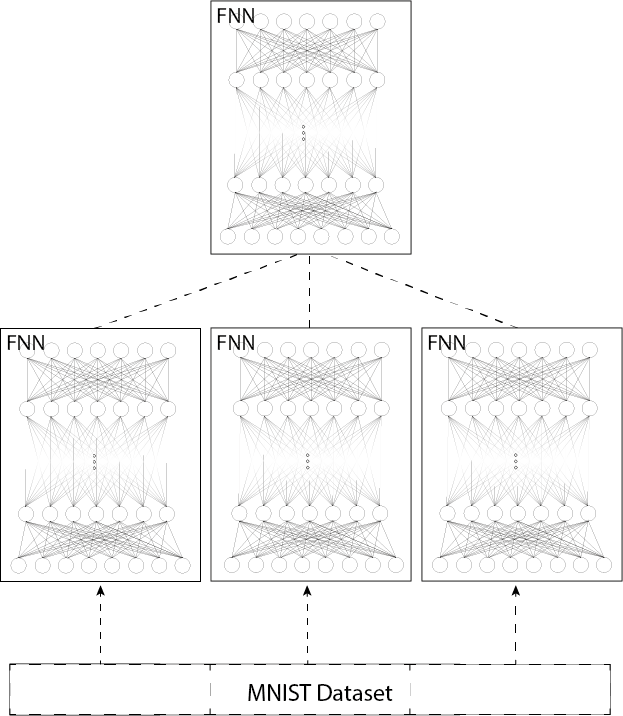}
\caption{Design of a Parallel Neural Network}
\label{fig:parallelDesign}
  \end{center}
\end{figure}

The parallel-design is visualized in figure~\ref{fig:parallelDesign}. On the bottom it shows
the dataset which is divided into as many slices as there are networks,
referred to as child-networks (CN). Each child-network learns only a slice of the dataset.
Ultimately the results of all parallel child-networks are merged to one final parallel neural network (PNN).
The combination of those CNs can be done in various ways. In the presented network the average of all
weights, calculated by each parallel CN by a set number of epochs, is used for the PNNs weights.
For the biases the same procedure is used, e.g. averaging all biases for the combined biases value.

In Golang it is important to take into consideration that a program, which
is designed parallel does not necessarily work in a parallel manner,
as a concurrent program can be parallel, but doesn't have to be.
This programming language offers a goroutine, which \qq{is a function executing concurrently
with other goroutines in the same address space} and processes with Go runtime.
To start a goroutine a $go func$ is called. It can an be equipped with a WaitGroup,
that ensures that the process does not finish until all running processes are done.
More about the implementation is explained in the next section.

\subsection{Implementation Details}

The main interface to which any trainable network binds is the $TrainableNetwork$ interface.
This interface is used throughout the whole learning and testing process.
Parallel - as well as simple neural networks implement this interface.
This allows for easy and interchangeable usage of both network types throughout
the code.
Due to the fact that a parallel neural network is built from multiple sequential neural networks (SNN)
we start with the implementation of an SNN.
The provided implementation of an SNN allows for a flexible network structure.
For example, the number of layers and neurons, as well as the activation-functions used on a layer,
can be chosen freely. All information, required for creating a network is stored
within a $NeuroConfig$ struct on a network instance. These settings can
easily be adjusted in a configuration file, the default name is $config.yaml$, located
in the same directory as the executable.

A network is built out of layers. A minimal network is at least composed of
an input layer and an output layer. Beyond this minimum, the hidden depth of
a network can be freely adjusted by providing a desired number of hidden layers.
Internally layers are represented by the $NeuroLayer$ struct.
A layer holds weights and biases which are represented by matrices. The Gonum
package is used to simplify the implementation. It provides a matrix implementation
as well as most necessary linear algebraic operations.

In the implementation, utility functions are provided for a convenient creation of
new layers with initialized weights and biases. The library $rand$ offers a
function $NormFloat64$, where the variance is set 1 and the mean 0 as default.
Weights are randomly generated using that normal distribution
seeded by the current time in nanoseconds.

The provided network supports several activation functions. The activation
function is defined on a per layer basis which enables the use of several
activations within one network.

A PNN is a combination of at least two SNN. The $ParallelNetwork$ struct
represents the PNN in the implementation. As SNNs are trained individually
before being combined with the output network of a PNN, it is necessary to
keep the references to the network managed in a slice. In the context of
a PNN the SNNs are referred to as child networks (CN).

In a PNN the training process is executed on all CNs in parallel using goroutines.
First, the dataset is split according to the amount of CNs.
Afterwards, the slices of the training dataset and CNs are called with
a goroutine. The goroutine executes minibatches of every CN in
parallel. Within those minibatches, another mutexed concurrent goroutine is
started for forwarding and backpropagating. Installing a mutex ensures
safe access of the data over multiple goroutines.

The last step of training is to combine those CNs to one PNN.
The provided
network uses as combination function the "average" approach. After training the CNs for a set number of
epochs, weights, and biases are added onto the PNN.
Ultimately these weights and biases are scaled by the
number of CNs. The result is the finished PNN.

\section{Performance Evaluation}\label{analysis}

At first a test with one PNN, consisting of 10 CNs, and
an SNN are tested using different activation functions on the hidden layer,
while always using the softmax function on the output layer. After
deciding on an activation function, network configurations are tested.
While the number of neurons is only an observation, but not thoroughly tested,
the number of networks is evaluated on different sized PNNs. Finally, the performance
of both types of networks are compared upon time, accuracy, confidence and costs.

\subsection{MNIST Dataset}
\label{section:mnist}

   For our analysis, we use the MNIST dataset which holds handwritten numbers and allows
    supervised learning. Using this dataset the network learns to
    read handwritten digits. Since learning is achieved by repeating a task,
    the MNIST dataset has a \qq{training-set of 60,000 examples, and a test-set
    of 10,000 examples}~\cite{mnist} .
    Each dataset is composed of an image-set and a label-set, which holds the information
    for the desired output and makes it possible to verify the networks output.
    All pictures are centered and uniform by 28x28 pixels. First, we
    start the training with the training-set. When the learning phase is over the network
    is supposed to be able to fulfill its task~\cite{liu2016accuracy}. To evaluate it's efficiency it is tested
    by running the neural network with the test-set since the samples of this set are still unknown.
    It is important to use foreign data to test a network since it is more
    qualified to show the generalization of a network and therefore its true efficiency.
    We are aware that MNIST is a rather small data set. However, it was chosen on purpose, because it is used in many similar parallelization approaches and allows therefore for relatively easy comparison of results.

    \subsection{Activation Functions in Single- and Parallel Neuronal Networks}

\begin{figure}[ht]
   \begin{center}
  \includegraphics[width=\textwidth]{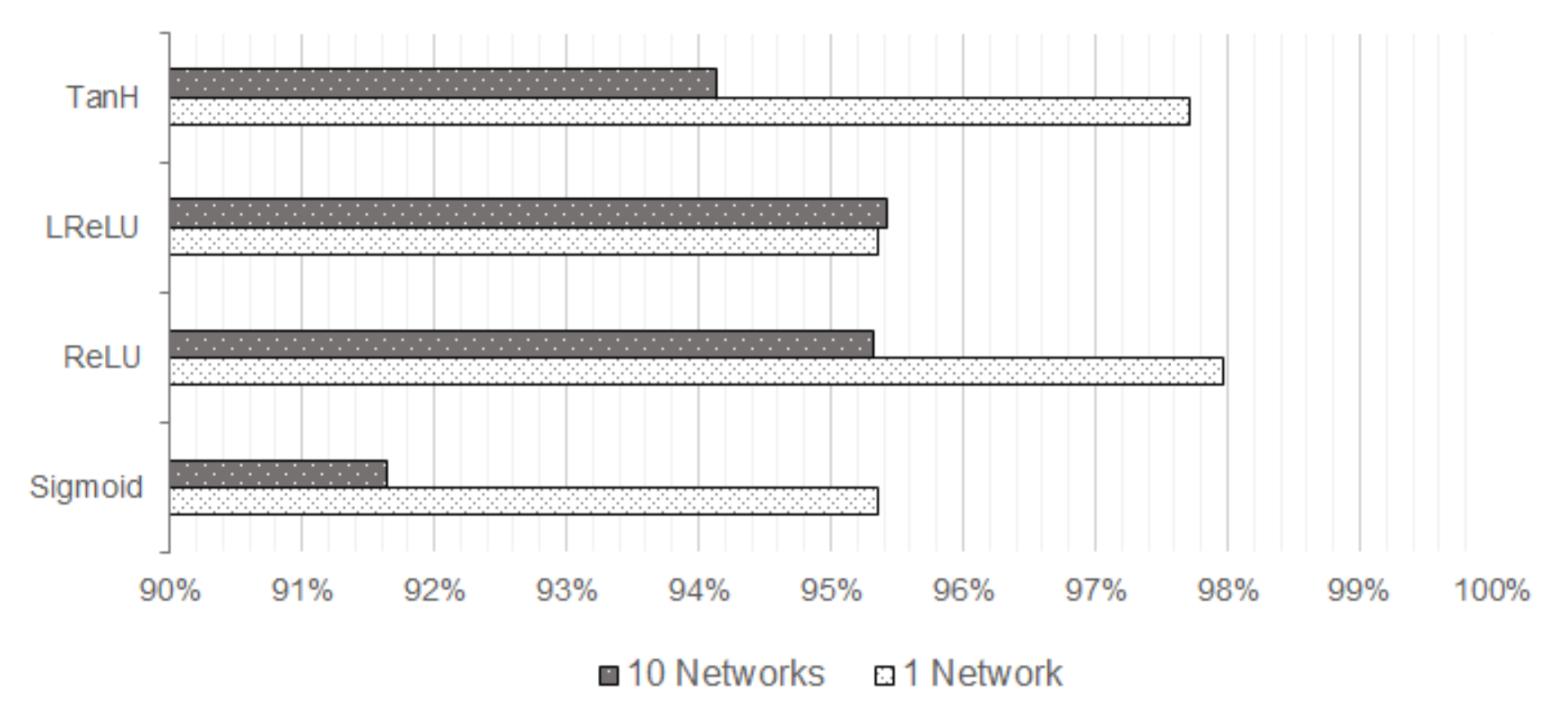}
  \caption{Compare Accuracy of a parallel- vs simple-NN with different activation functions and a softmax function for the output layer.
  The networks have one hidden layer with 256 neurons and the training was performed with a
  learningrate of 0.05 and a batchsize of 50 over 20 epochs.}
  \label{fig:activationFuncs}
  \end{center}
\end{figure}

  To elaborate which function performes best in terms of accuracy for the coded single- and parallel neural
  network a test using the same network design and settings for each network is performed while changing
  only the function used on the hidden layer.
  Used settings were one hidden layer built out of 256 neurons, working with a batchsize of 50 and a
  learningrate $\eta$ of 0.05 and an output layer calculated with softmax. This is used on a single FNN and a
  PNN each consisting of 10 child-networks.
    Figure~\ref{fig:activationFuncs} presents the performance results of the activation functions.
  Each networks setup is one hidden layer on which either the tangent hyperbolic-, leaky ReLU-, ReLU-
  or sigmoid-function was applied.

  In this comparison the
  single neural network that learned using the ReLU-function, closely followed by TanH-function,
  has reached the best result within 20 epochs. While testing different configurations it showed that most
  activation functions reached higher accuracy when using small learning rates. Sigmoid is one
  function that proved itself to be most efficient when the learningrate is not too small. By raising
  the learningrate to 0.6 the sigmoid-functions merit grows significantly on both network types.
  In the process of testing ReLU on hidden layers in combination with Softmax for the output layer has
  proven to reliably deliver good results. That is why in further sections
  ReLU has applied on all networks hidden layers and on the output layer Softmax.

	\subsection{Network Configurations}
  \label{subsection:netConfigs}

    \subsubsection{Number of Neurons.}
    Choosing an efficient number of neurons is important, but it is hard to identify.
    There is no calculation which helps to define an effectively working number or range
    of neurons for a certain configuration of a neural network.
    Varying the number of neurons between 20 to 600 delivered great accuracy.
    These are only observations and need to be studied with a more sophisticated approach.

    \subsubsection{Number of Networks.}
\begin{figure}[ht]
      \begin{center}
      \includegraphics[width=12 cm]{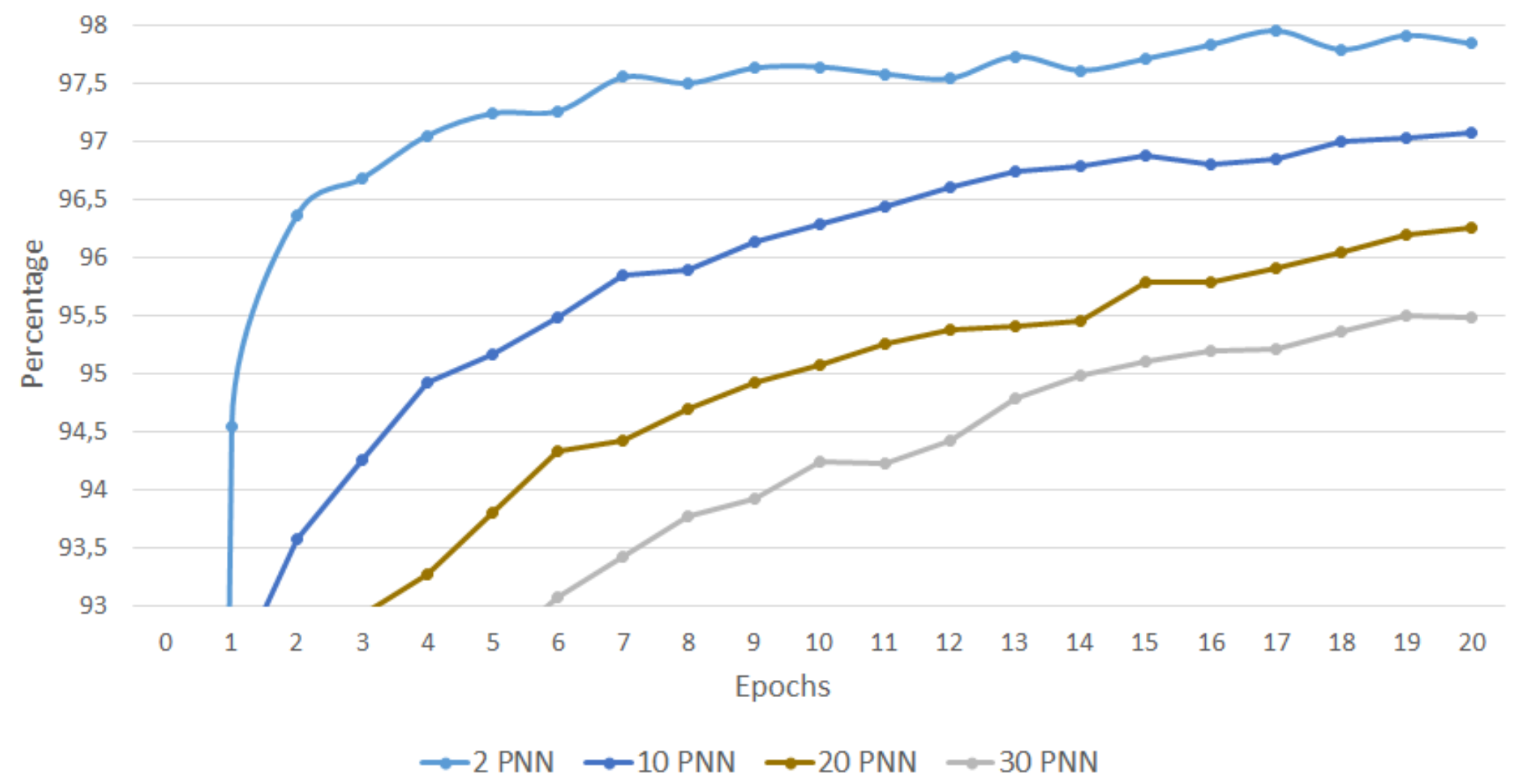}
      \caption{Accuracy of PNNs, built with different amount of CNs, over 20 epochs }
      \label{fig:parAve}
      \end{center}
      \end{figure}

    To evaluate the performance of PNNs in terms of accuracy,
    PNNs with different amounts of CNs are composed and trained. The training runs
    over 20 epochs with a learning rate of 0.1 and a batchsize of 50.
    All CNs are built with one hidden layer consisting of 256 neurons. On the hidden layer the ReLU-function and
    on the output layer the Softmax-function is used. After every epoch, the networks are tested with
    the test-dataset. The results are visualized in figure~\ref{fig:parAve}.

Figure~\ref{fig:parAve} illustrates a clear loss in accuracy of PNNs with a growing number
of CNs. The 94.5\% accuracy, for example, is reached by a PNN with 2 CNs after only one epoch,
while a PNN with 30 CNs achieves that after 12 epochs.
In respect to the number of networks this graph shows that more is not always better.
Considering, that this test was only performed over a small number of epochs, it is not
possible to read the potential of a PNN with more CNs. To find out how good a PNN
can perform, a test was run with three PNNs running 300 epochs:

\begin{table}
  \centering
	\begin{tabular}{ | c | c | c | c | c | c | c | } \hline
    	\multicolumn{1}{|c|}{CNs of PNN} &  \multicolumn{6}{c|}{Accuracy after...}\\  \cline{2-7}
        & 20 Epochs & 100 Epochs & 150 Epochs & 200 Epochs & 250 Epochs & 300 Epochs \\ \hline
        2 & 97.76 & 98.08 & 98.13 & 98.16 & 98.14 & 98.17  \\ \hline
    	10 & 96.58 & 97.43 & 97.96 & 98.03 & 98.09 & 98.05  \\ \hline
   	 	20 & 95.69 & 97.50 & 97.71 & 97.92 & 97.90 &97.97 \\ \hline

  \end{tabular}
  \caption{Accuracy behaviour for different epochs}

  \label{tabular:300}
\end{table}

\begin{figure}[ht]
  \begin{center}
    \includegraphics[width=\textwidth]{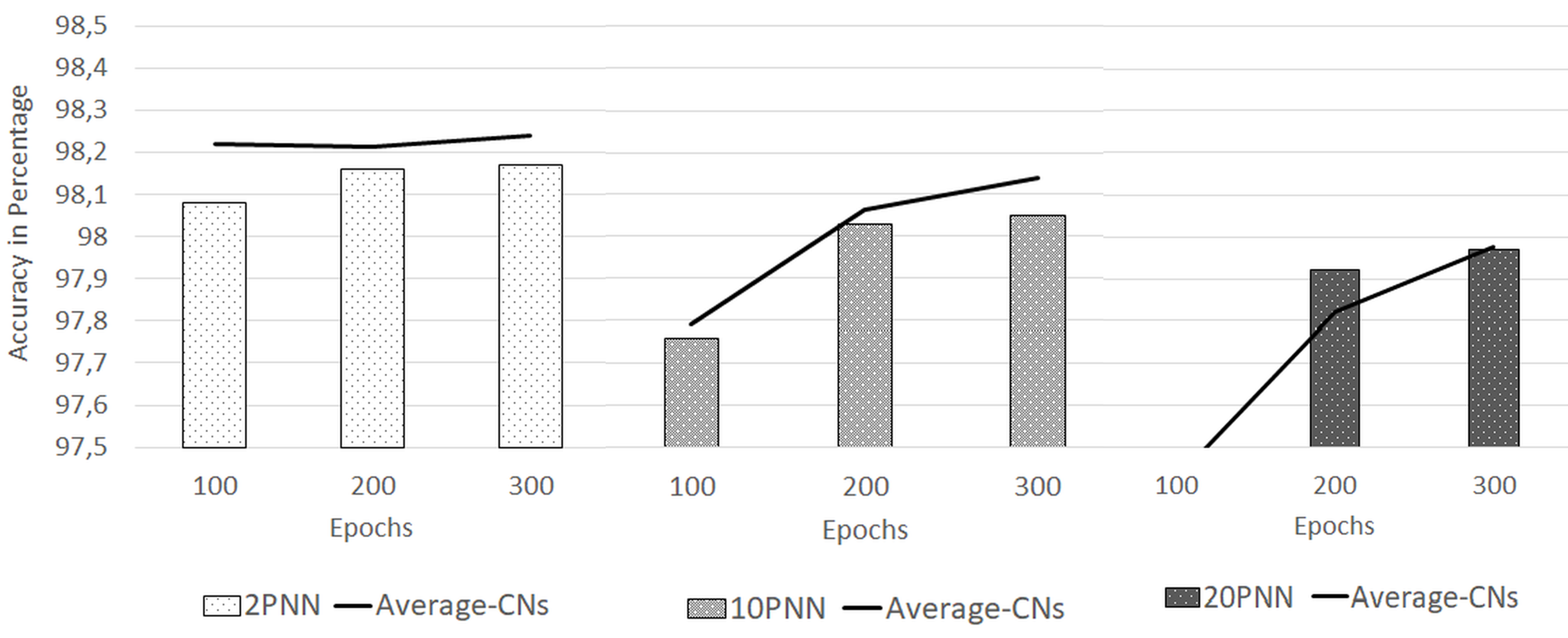}
  \end{center}
    \caption{Compare the average accuracy of all CNs, out of which the final PNN is formed, with that PNNs
    accuracy}
    \label{fig:CNvsPNN}
\end{figure}

Table~\ref{tabular:300} shows a static growth until 200 epochs. After that, there is only a small
fluctuation of accuracy, showing that a local minimum has been reached.
Over the runtime of 300 epochs the difference of the performance regarding the accuracy
of PNNs has been reduced significantly. Still the observation of the ranking of the PNNs
has not been changed. The PNNs built out of a smaller number of CNs perform slightly
better.
Since the provided PNNs are built by using averaging of weights and biases
it also seemed interesting to compare the average accuracy of the CNs
with the resulting PNN, to grade the used combination function.
The results are illustrated in figure~\ref{fig:CNvsPNN}.

It shows that the efficiency of an average function grows with the number
of CNs. The first graph drawn with 2 CNs shows, that the resulting PNN is performing
worse than the average of the CNs, it has been built from. By growing the number of
CNs to 10, the average of CNs approximates towards the PNN. The last graph of this figure
shows that a PNN composed of 20 CNs outperforms the average of its CNs after 200 epochs, and
after 300 epochs levels with it.
It has to be noted that the differences in accuracy are very small, as it is only
a range of 0.1 to 0.2 percent. Overall it can be said that this combination function
is working efficiently.

    \subsection{Comparing the Performances}

    \subsubsection{Time.}


    Time is the main reason
    to have a network working in parallel.
    To test the effect of parallelism on the time required to train a PNN,
    the provided neuronal network is tested on three systems. The first system
    is equipped with 4 physical and 4 logical cores, an Intel i7-3635QM processor working
    with a basic clock rate of 2.4GHz, the second system holds 6 physical cores and 6 logical cores
    working with 2.9GHz and an Intel i9-8950HK processor and
    last the third system works with an AMD Ryzen Threadripper 1950X with 16 physical and 16 logical cores,
    which work with a clock rate of 3.4GHz. The first, second and third systems are referred to as 4 core, 6 core
    and 16 core in the following.

    \begin{figure}[ht]
      \begin{center}
      \includegraphics[width=11cm]{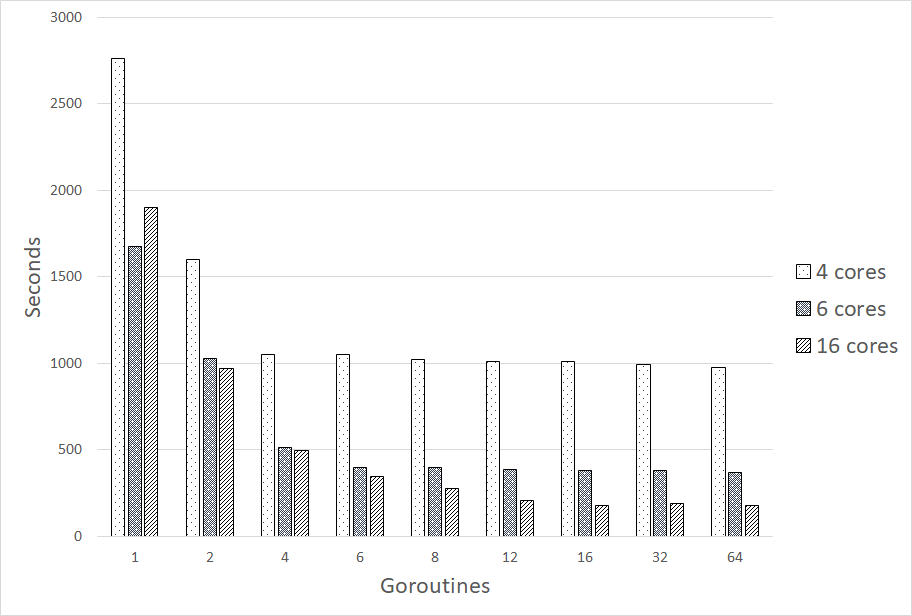}
      \caption{Time in seconds, that was needed to train a PNN with a limited amount of
      one Goroutine per composed CN.}
      \label{coreSpeed}
      \end{center}
    \end{figure}

    In figure~\ref{coreSpeed} the benefit in terms of time using
    parallelism is clearly visible.
    The results illustrated show the average time in seconds needed by each system
    for training a PNN consisting of one CN per goroutine. For the block diagram in~\ref{coreSpeed} the percental time requirements
    in comparison with the time needed using one goroutine are listed in table~\ref{tabular:coreSpeed}.

    \begin{table}
      \centering
      \begin{tabular}{ | c | c | c | c | c | c | c | c | c | c | } \hline
          \multicolumn{1}{|c|}{} &  \multicolumn{9}{c|}{Time required compared to 1 goroutine}\\  \cline{1-10}
          System/Goroutines & 1 & 2 & 4 & 6 & 8 & 12 & 16 & 32 & 64 \\ \hline
          4 & 100\% & 58\%	& 38\% &	38\% &	37\% &	37\% &	37\% &	36\% &	35\%
            \\ \hline
          6 & 100\% & 61\% &	31\% &	24\% &	24\% &	23\% &	23\% &	23\% &	22\%
          \\ \hline
            16 & 100\% & 51\% &	26\% &	18\% &	14\% &	11\% &	9\% &	10\% &	9\%
            \\ \hline

      \end{tabular}
      \caption{Average time required to train a PNN in comparison to one goroutine,
      which represents 100 percent}
      \label{tabular:coreSpeed}
    \end{table}

    The time in figure~\ref{coreSpeed} starts on a high level and decreases with an increasing amount
    of goroutines for all three systems. Especially in the range of 1 to 4 goroutines, a formidable decrease in training time is visible
    and only starts to level out when reaching a systems physical core limitation.
    This means that the 4 core starts to level out after 4 goroutines, the 6 core after
    6 goroutines and the 16 core after 16 goroutines, even though all systems support
    hyper threading. After reaching a systems core number the average time necessary for
    training a neural network decreases further with more goroutines. This should be
    due to the ability to work in parallel and in concurrency as one slot finishes
    and a waiting thread can start running immediately, without waiting for the rest of the running
    threads to be finished.
    All three systems show high time savings by parallelizing the neural networks. While time requirements
    decreased in every system, the actual time savings differ greatly as
    the 16 core system decreased 91 percent on average from 1 goroutine to 64 goroutines. In comparison,
    the 4 core system only took 65 percent less time. As the 16 core system is a
    lot more powerful than the 4 core system, it can perform an even greater parallel task and therefore displays
    a positive effect of parallelism upon time requirements.
    Based upon figure~\ref{coreSpeed} and its table~\ref{tabular:coreSpeed} parallelism within neural networks can
    be seen as a useful feature.

\subsubsection{Accuracy and Confidence of Networks.}

\begin{figure}[ht]
  \begin{center}
  \includegraphics[width=0.8\textwidth]{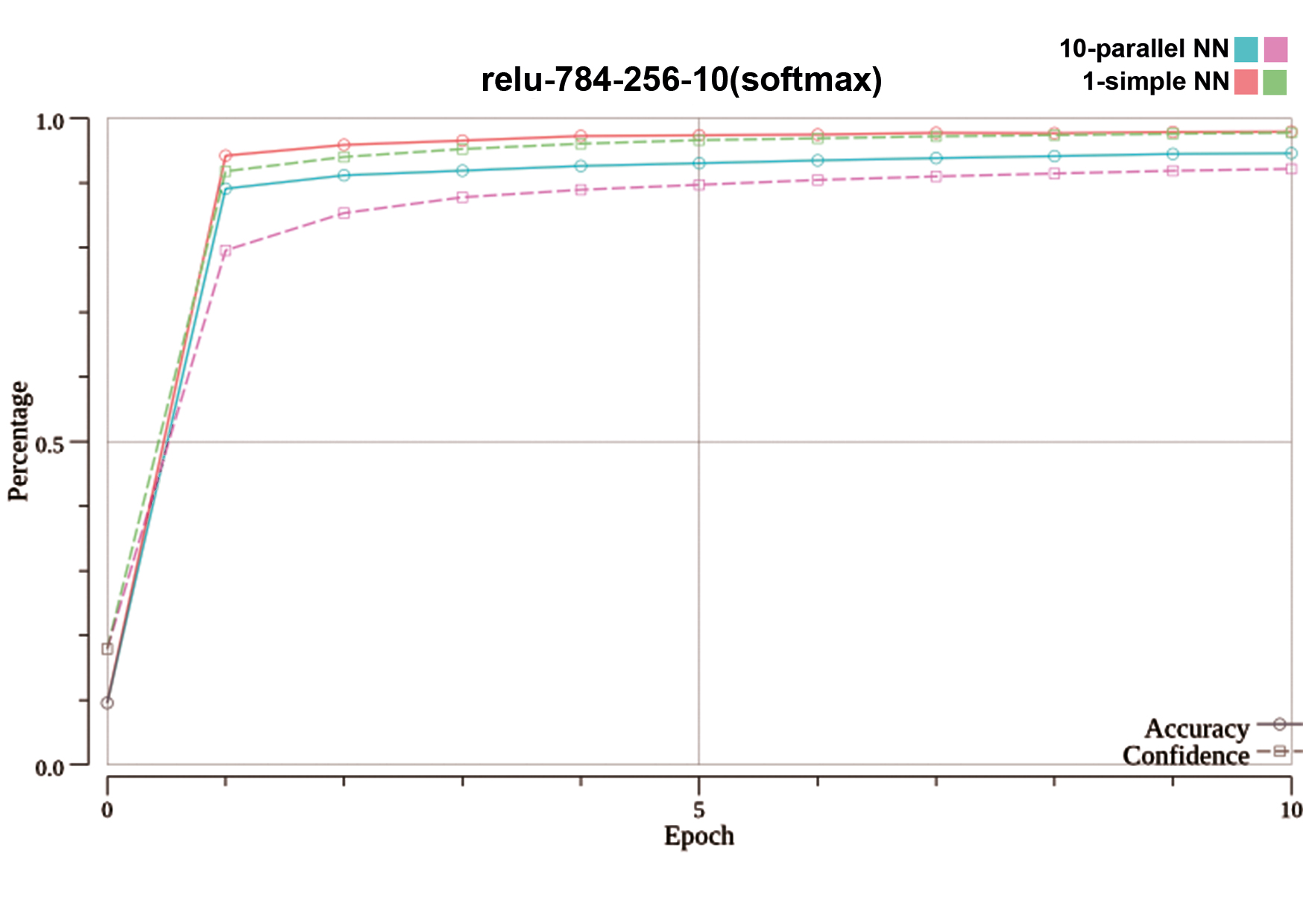}
  \caption{Compare Accuracy and Confidence of a PNN composed of 10 CNs and an SNN with one Hidden Layer which holds 256 Neurons}
  \label{performanceParVsSin}
  \end{center}
\end{figure}

 In this section the performance in terms of accuracy and confidence is compared
 between a PNNs and an SNN.
 For the test, illustrated by figure~\ref{performanceParVsSin}, both types of networks have been
 provided with the same random network to start their training. They have the exact same
 built, except that one is trained as SNN and the other is cloned 10 times to build a
 PNN with 10 CNs.

In figure~\ref{performanceParVsSin} the SNN performs better than
the PNN in both accuracy and confidence. While the SNNs accuracy and confidence
overlap after 8 epochs, the PNN has a gap between both lines at all times.
This concludes that the SNN is "sure" about its outputs, while the PNN is more volatile.
The SNNs curve of confidence is a lot steeper
than the PNNs and quickly approximates towards the curve of accuracy.
Both curves of accuracy start off almost symmetric upwards the y-axis, but the PNN
levels horizontally after about 90 percent while the SNN still rises until about
94 percent. After those points both accuracy curves run almost horizontally and in parallel
towards the x-axis. The gap stays constantly until the end of the test.
Even small changes within the range of 90 to 100 percent are to be interpreted as
significant. This makes the SNN perform considerable more efficient in terms
of accuracy and costs than the PNN.

\subsubsection{Cost of Networks.}
\begin{figure}[ht]
  \begin{center}
    \includegraphics[width=12 cm]{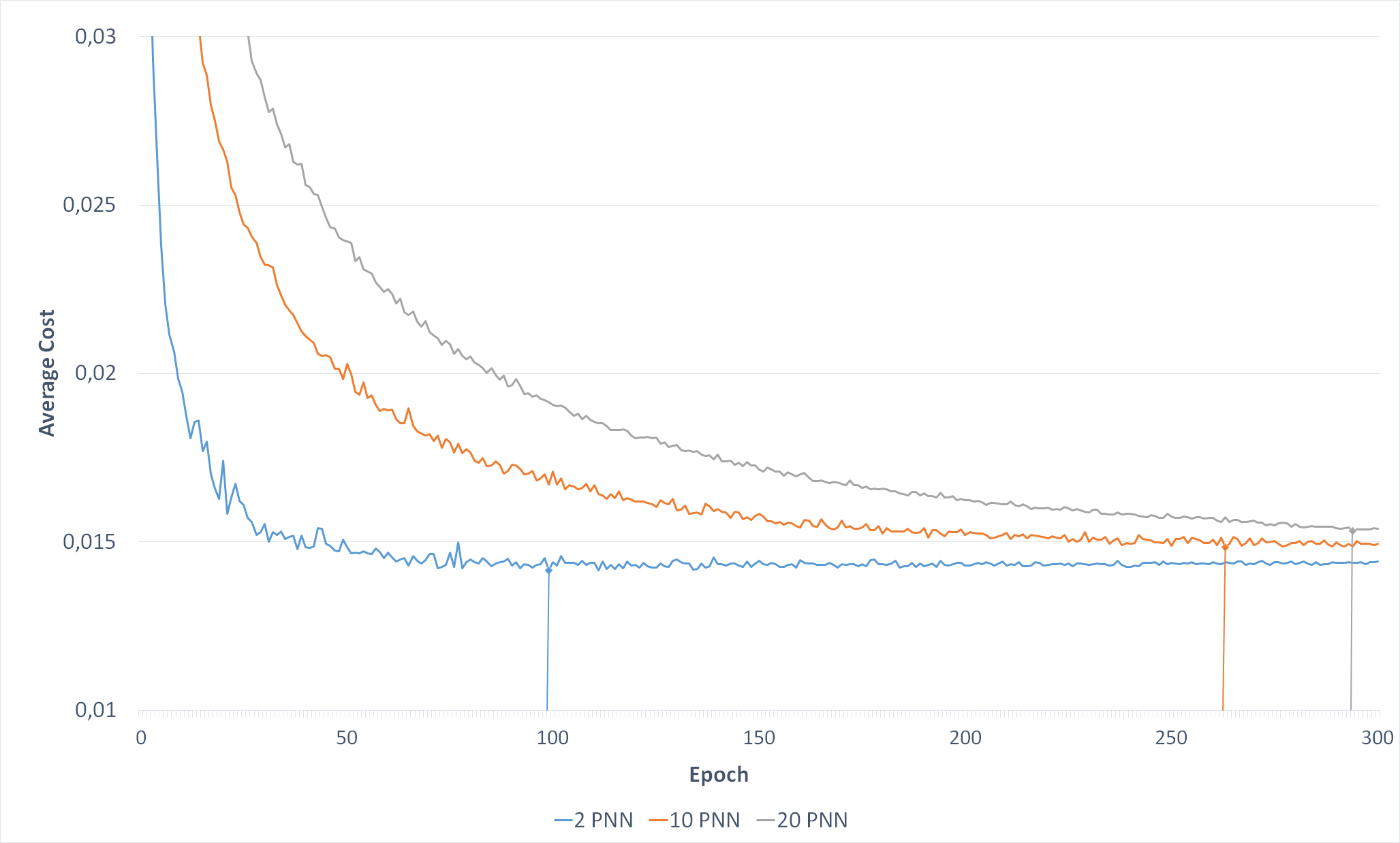}
    \caption{Average Costs of PNNs over 300 epochs. The vertical
    lines show the lowest cost for each PNN.}
    \label{fig:Cost}
  \end{center}
\end{figure}

To see how successful the training of different PNNs are,
the costs of 3 parallel networks with a varying number of CNs have been recorded for 300 epochs. The results are
illustrated in figure~\ref{fig:Cost}.

It shows that the costs of all three PNNs sink rapidly within the first 50 epochs.
Afterwards, the error decreases slower, drawing a soft curve that flats out towards
a line, almost stagnating. Apparently, all PNNs training moves fast towards a minimum at the beginning,
then slows down and finally gets stuck, while only moving slightly up and down the minimums borders.
Similar to earlier tests, a PNN built with less CNs performs better. More CNs leave the graph
further up the y-axis, as the 2-PNN outperforms both the 10- and 20-PNN. It also reaches its
best configuration, e.g. the point where costs are lowest, significantly earlier than the
other tested PNNs. Whereas the 10- and 20-PNNs work out their best performance regarding the costs
at a relatively close range of epochs, they reach it late compared to the 2-PNN.
Figure~\ref{fig:Cost} clearly shows a decrease in quality with PNNs, formed with more CNs. This indicates
that the combination function needs optimization to achieve a better graph. In the long term,F costs behave the same
as accuracy. After 300 epochs the difference has almost leveled.

    \section{Findings and Conclusion}

    This paper presents and analyses PNNs composed of several
    sequential neural networks.
    The PNNs are tested upon time, accuracy and costs and compared to an SNN.

    The parallelization approach on three different multicore systems show excellent speedup (the time necessary for training a PNN reduces constantly
    by increasing the number of CNs e.g. number of goroutines).

%

    With all three tested systems the time necessary for training a PNN decreased constantly
    by increasing the number of CNs e.g. number of goroutines.
    While the difference in time was significant within the first few added goroutines
    it leveled out after reaching the systems number of cores.
    A PNN with 2 CNs takes 40\% to 50\% less time than a SNN and a PNN with 4 CNs takes 60\% to 70\% less time.

    While time is a strong point of the PNN, accuracy is also dependent on the number
    of CNs a PNN is formed from. While a few CNs resulted in longer training times it generated better accuracy in fewer epochs.
    More CNs made the training time faster but the learning process slower. After 20 epochs a PNN composed of 2 CNs
    reached an accuracy of almost 98\%, while a PNN composed of 20 CNs only slightly
    overcame the 96\% line. When both PNNs were trained for a longer period this difference shrank dramatically.
    Trained for 300 epochs the accuracy only differed by 0.2\% in favor of the
    PNN made out of 2 CNs. While this proved the ability to learn with a small data set it also
    demonstrated that bigger data sets deliver a better result faster. 
    the PNNs can improve by 0.41\% and 2.28\% when training for a longer period.
    These results were achieved by using averaging as combination function. The chances of achieving an even better
    accuracy by improving the combination function are high. The costs of a PNN also depends on the number of CNs. It has the same behavior as accuracy and can
    also be improved by an optimized combination function. However, a thorough analysis on the effects of improved combination functions is planned for future work
    and is beyond the scope of this paper.


    Summing up, PNNs proved to be very time efficient but are still lacking in terms of accuracy. As there are plenty of
    other optimizations, e.g. adjusting learning rates~\cite{DBLP:journals/corr/GoyalDGNWKTJH17},
a PNN  proved to be more time efficient than an SNN. However, until the
    issue of accuracy has been taken care of, the SNN surpasses the PNN in practice.

    We close the paper with a final word on the feasibility of Golang for parallel neural network simulation:
    Data parallelism proved to be an efficient parallelization strategy.
    In combination with the programming language Go, a parallel neural network implementation is coded as fast as a sequential one, as no special efforts are necessary for concurrent programming thanks to Go’s concurrency primitives, which offer
a simple solution for multithreading.

%

\bibliographystyle{splncs04}
\bibliography{references}

\end{document}